# An Evolved Neural Controller for Bipedal Walking with Dynamic Balance

Michael E. Palmer
1859 Laboratories, Inc.
922 Matadero Ave.
Palo Alto, CA 94306
+1-415-867-3653

mep@1859labs.com

Daniel B. Miller
Singular Robotics
1515 Channing Ave.
Palo Alto, CA 94303
+1-650-815-8801

danbmil99@gmail.com

## ABSTRACT
We successfully evolved a neural network controller that produces dynamic walking in a simulated bipedal robot with compliant actuators, a difficult control problem. The evolutionary evaluation uses a detailed software simulation of a physical robot. We describe: 1) a novel theoretical method to encourage populations to evolve "around" local optima, which employs multiple demes and fitness functions of progressively increasing difficulty, and 2) the novel genetic representation of the neural controller.

## Categories and Subject Descriptors
D.2.2 [**Software Engineering**]: Evolutionary Prototyping; I.2.8 [**Artificial Intelligence**]: Problem Solving, Control Methods, and Search – *heuristic methods*; I.2.8 [**Artificial Intelligence**]: Distributed Artificial Intelligence.

## 1. INTRODUCTION
General control of legged robots for dynamic walking remains an open problem. Several authors have used neural networks with either evolved or learned weights to perform bipedal control [1-5]. Dynamic walking, wherein the robot momentarily falls at points during the walk, is more difficult than static walking, in which the robot is continuously supported by the (typically larger) support envelope of the robot's feet. Use of compliant actuators, e.g., [6], rather than stiff actuators, further increases the difficulty of the control problem.

This work successfully addressed the problem of evolving a neural network controller to make a compliant bipedal robot balance, walk dynamically forward, and balance again in simulation. There are several aspects of interest:

We provide the robot with several partially contradictory objectives to optimize. First, we have a parametric description of the optimal movement of the robot's center of mass (COM) over time. Second, we impose constraints on the roll, pitch, and yaw of the robot's torso. Third, we provide a parametric "suggested" trajectory for all joints of the robot, roughly resembling a human walking motion, which is computed with simple sine functions. These objectives are partially contradictory: for example, if the suggested joint trajectories are followed exactly, the robot will not balance, nor will its COM exactly follow the suggested path. In addition, at different points along the walking path, there are different compromises to be made among these objectives.

Responsibility for control of the robot is divided among multiple independent neural networks. Each network controls the robot for a short time. Each is assigned to a single chromosome.

We describe novel methods of manipulating population structure to encourage evolved populations of neural controllers toward completion of the entire walking task, and away from becoming trapped in local optima. Specifically, we create a set of 19 demes arranged linearly in space. Migration is permitted between neighboring demes. The fitness task becomes increasingly challenging as individuals migrate from left to right through the demes, and individuals are prevented from invading empty demes before they have attained a threshold fitness value. Nonetheless, empty demes present an opportunity for the individuals that are able to invade.

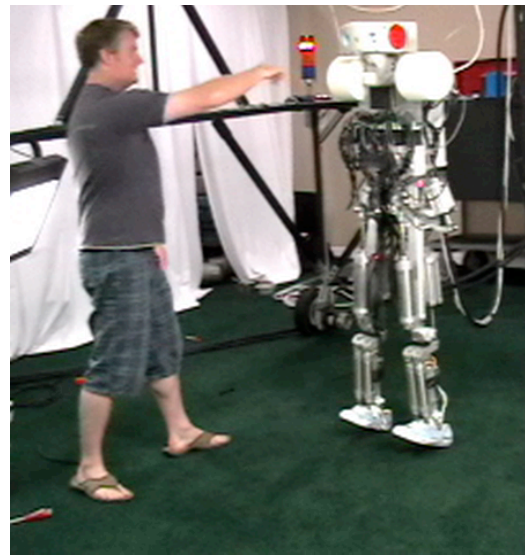

**Figure 1: Anybots' Dexter hardware, on right. See [7]. A biologically-evolved bipedal walker is on left.**

## 2. METHODS
### 2.1 Robot Hardware

This work uses a detailed simulation of a hardware robot called Dexter, a 12-degree-of-freedom (DOF) bipedal walking robot developed by Anybots, Inc. The Dexter hardware is shown in Figure 1. Dexter has pneumatic actuators, as opposed to more typical electromechanical or hydraulic actuators. Pneumatic actuators have a high level of compliance, or low stiffness, compared to other actuators. This high compliance results in a significantly more difficult control problem, but reduces the likelihood of damage to the robot or humans in the event of a fall or other malfunction. Compliant actuation also allows energy capture and release, leading to greater energy efficiency. Therefore there is interest in the robotics community in control strategies that work on robots with compliant actuators. Creating a successful controller for a walking 12-DOF bipedal robot with compliant actuators is a challenging, open problem. With compliant actuation, unlike with stiff actuation, analysis paradigms such as Zero Moment Point (ZMP) [8] offer only limited help. Position-based control of each joint is not nearly precise enough to develop a stable walking gait using positional feedback alone, as in the traditional ZMP approach. In addition, changes to any single actuator's input can have a significant impact on the future position of other actuators.

Each of the 12 joints has a sensor that detects the current joint angle. In addition, a 6 DOF inertial measurement unit (IMU) is mounted on the torso of the robot, providing data on X, Y, and Z accelerations; and pitch, roll, and yaw rates.

### 2.2 Physical Simulator

Anybots also developed simulation software to model Dexter, written in C++ and Python, utilizing the Open Dynamics Engine, an open-source rigid body simulation system. The simulation was developed specifically to model the complexities of pneumatic control. In particular, a thermodynamic model of temperature and pressure changes over time was incorporated into the actuator model; the parameters of this model were tuned to match the empirical response of the actual robot. The simulated version of Dexter is illustrated in Figure 2.

From the point of view of the control software, the simulated robot can be summarized as a "black box" with 12 inputs (the 12 joint actuators) and 18 outputs (the 12 joint angles, plus the 6 degrees of freedom of the IMU). Every 10ms, a neural network controller computes a new set of the 12 actuator values; the physical simulator computes the positions of all parts of the robot for the next time step given those actuator motions; and a subset of the sensor values are processed (as described below) and passed to the controller to compute its next actions.

### 2.3 Walking Task and Suggested Trajectories

The walking task is as follows: starting from a balance phase, we desire the robot to take five steps, resulting in a forward translation. The robot should then revert to its balance mode, and remain standing upright. The entire walking task, not including the balancing phases, is 3.4 seconds long.

The search space of motions of all actuators to move the robot from the starting point to the end is multidimensional and very large. Most of the state space will not produce anything nearly resembling a human walk. We greatly constrain the search by providing a set of target or "suggested" trajectories of three types, parameterized in time, for the robot to follow.

First, we generated a set of suggested trajectories for the angles of all joints. This set was generated algorithmically, using sine-based generators, and resembles a basic human walking cycle, including an initial half-step of the right leg, followed by three full steps of the left, right, and left legs, and a final half-step of the right leg, bringing the robot back to a standing pose. Note that if the robot were to follow the suggested joint angles exactly, it would fail to balance. However, as it turns out, we were indeed able to evolve a successful controller that balances throughout the walking task, and produces output not very distant in state space from the suggested joint trajectories.

In addition, we provide a suggested trajectory for the center of mass of the robot, consisting of an acceleration phase, coasting phase, and deceleration phase. The overall desired forward translation of the center of mass is approximately 1.2 meters.

Finally, our desired trajectory should minimize variations in the pitch, roll, and yaw of the torso; we want the torso to experience a smooth ride.

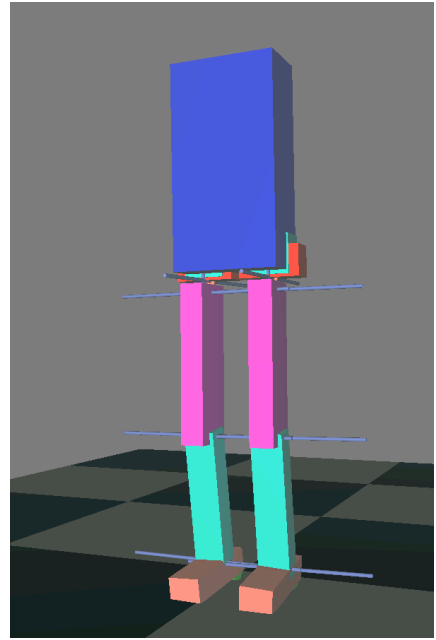

**Figure 2: Simulated version of Dexter. See [9].**

Note that the above suggestions do not include information to assist the robot in maintaining its balance. The success of the task depends on the control networks deviating from the suggested trajectories as necessary to effect foot placement and ankle torque (as well as hip torque when both feet are in contact with the ground), in order to constantly maintain balance throughout the walking task.

### 2.4 One Evolutionary Individual: a Set of Neural Networks

We divided the walking task of 3.4 seconds duration into 34 time slices of 100 milliseconds each. The phenotype of one evolutionary individual consists of 22 separate neural network controllers. Each of these networks is responsible for control of the robot for one or more of the 34 time slices (some networks are reused for multiple time slices, as detailed below). In addition, there is one pre-evolved network that is responsible for standing balance only; this network is identical for all individuals, does not evolve, and is in control at the beginning and end of the walking task when the robot is (ideally) standing still and upright.

To evaluate the performance of a given set of neural networks, we run a series of tests using the networks to control the simulated robot. Each test is identical, except that we slightly vary the initial conditions to ensure that the simulator does not repeat its previous run precisely. The randomization consists of applying a random force impulse to the robot, in a randomized direction, while balancing, before it begins to walk. In addition, we vary the distance between the feet, on the order of several centimeters.

During each test, the simulated robot is lowered to the ground, and the pre-evolved balancing network is activated. We allow the robot to balance for several seconds. Then, we transition from balancing to the walking phase. Each neural network has primary control of the robot for 100 milliseconds at a time. However, straddling the points of transition between networks, we apply a 50ms period of "cross-fade" between networks, to smooth the transitions from one to the next. During this cross-fade period, the earlier network's weight ramps down from 1.0 to 0.0, while the weighting of the next network ramps up from 0.0 to 1.0. This pattern continues until the end of the walk, at which point we transition back to the balancing network.

Most attempts at the task result in the robot losing its balance before completion of the entire walking task. (E.g., see Figure 7, rows 1 and 2. The figure is discussed in more detail below.) As evolution progresses and the individuals in the population improve, the average individual is able to progress further toward completion of the entire task before loss of balance. (See Figure 7, rows 3, 4, and 5.)

## 2.5 Neural Networks

The 22 neural networks in the phenotype of each individual are given one of four different topologies. In the work discussed here, the edge weights, but not the topologies, evolve. The four topologies are called: *balance*, *step-off*, *walk-left*, and *walk-right*. The *walk-right* topology is illustrated in Figure 3. The sensor names correspond to measurements of the torso orientation and velocity; the effector names correspond to the various actuators on the legs. The *walk-left* topology is identical, except the *Lean Y*, *Velocity Y*, and *Yaw* input signals are negated, and the outputs are reversed between the left and right legs, to exploit the bilateral symmetry of the robot (see below). The *step-off* topology (not shown) is used to initiate the walk, and has four inputs and eight outputs; the *Yaw* sensor is not connected, and the *Inner* and *Outer* effectors of each ankle are connected from the same output. The *balance* topology (not shown), used at the beginning and end of the walking task, is very simple, containing only 5 internal neurons; only the *Lean Y* and *Velocity Y* sensors are connected because the robot is intrinsically stable from side to side. We had prior domain knowledge indicating that this simple topology is sufficient to produce balance when the robot is standing. A single network of the *balance* topology was pre-evolved to balance the robot in a standing position; it does not evolve in the experiments here.

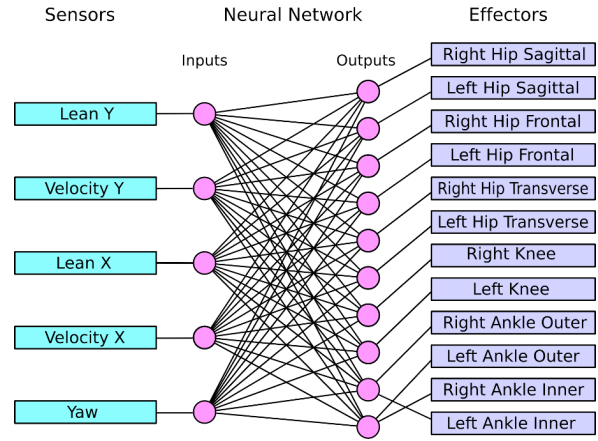

**Figure 3: The *walk-right* topology type**

To produce the final output that is sent to the joint actuators, the output neuron values, which may be positive or negative, are added to the suggested joint trajectories. In other words, the network outputs represent deviations from the suggested trajectories, computed in response to the inputs. Each actuator input indicates a nominal position for that actuator, specifying the position the joint would settle to with no outside forces acting on it. In practice, the actual actuator position deviates from the input value depending on how it is loaded; this can be modeled as an absolute position control with an attached spring.

At the beginning of a test, a single pre-evolved network of the *balance* topology runs for several seconds as the standing robot settles; this period is not counted in scoring. Then, a sequence of four networks of the *step-off* topology run for 100ms each; this sequence permits a lean to the left, to allow the robot to subsequently lift its right foot. Then six *walk-right* networks are used for acceleration up to normal walking speed. In the middle of the walk, another set of six networks is reused three times: first this set is connected as *walk-left*; then it is reused but connected as *walk-right*; then it is reused again, connected as *walk-left*. (This reuse exploits the bilateral symmetry of the robot; note the corresponding bilateral symmetry of the *walk-left* and *walk-right* topologies.) Finally, a sequence of six networks connected in topology *walk-right* is run as the robot decelerates to a stop; and the pre-evolved balance network is again activated. This yields a total of 22 (4 stepping off + 6 accelerating + 6 walking at constant speed + 6 decelerating) evolved networks, walking for a total of 3.4 seconds ((4 + 6 + 18 + 6) * 100ms = 3.4 seconds). One of the features enabled by this scheme is that we may naturally extend the middle portion of the walk (the six networks that are reused as *walk-left*, *walk-right*, *walk-left*) to a greater length.

## 2.6 Scoring

At the end of each test, we compute a score, intended to represent how well the individual (i.e., the set of 22 networks) performed for that test. The score is a weighted sum of several metrics. The predominant metric is time, which has a positive weight. The rest of the metrics are errors that are weighted negatively. These metrics are calculated as the sum of either the absolute value or the square of an error term for each 10-millisecond simulation step of the test. The metrics used are: variation in pitch, roll, and yaw from an upright position; the difference between the robot's actual COM and its suggested

COM trajectory; the difference between the robot's actual and target joint angles (lightly weighted); and the difference (in projected X, Y coordinates) between the center of mass of the torso and the mean position of the center of both feet.

Each individual makes five attempts at the walking task, and the five scores are averaged to produce a final score.

## 2.7 Multiple Demes and Progressive Fitness Functions

Here, we describe the novel theoretical aspect of this work. Gavrilets [10] describes how high-dimensional fitness landscapes differ qualitatively from low-dimensional ones: the former are likely to contain "extra-dimensional bypasses" that allow neutral evolution across "fitness valleys" that exist in lower-dimensional subspaces. The higher the dimensionality of the fitness landscape, the easier it is for genotypes to spread neutrally along "ridges" through all of genotype space, given a fixed probability of viability for any point in genotype space. Gavrilets considered idealized fitness landscapes with just two fitness values: viable, and inviable. In contrast, practical fitness landscapes have a range of fitnesses, complicating the picture beyond Gavrilets' original analysis. A barrier of inviable genotypes is not required to create a fitness valley in practice; all that is required is a barrier of sufficiently lower-fitness genotypes (where "sufficiently" depends on population parameters).

Figure 4 illustrates a population becoming trapped in a local optimum. Fitness is shown on the Y axis, and progress through the walking task on the X axis. Assume that the experimenter is unable to create a fitness function that increases monotonically in fitness as "progress" through the walking task is made, because of the complexity of the task (e.g., such was the case with our multi-objective walking task defined above). In the figure, the population starts at the point labeled 1 and is able to "hill climb" (that is, if mutation produces small movements on the X axis, then evolution will cause fitness to increase) easily through label 2 to label 3. Here, unfortunately, any small genotypic change creates only a decrease in fitness, so the population becomes trapped, unable to evolve solutions that progress further through the walking task, even though globally better walks exist.

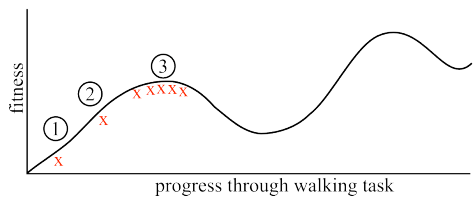

**Figure 4: One deme: trapped in a local optimum**

Gavrilets' model suggested to us that a population could be caused to neutrally explore more of genotype space if a portion of the fitness function were ignored, which would create larger networks of neutral ridges in genotype space. Certainly, though, if continuing adaptive progress is desired, selection must be present sometime or somewhere; hence, we may create neutrality either at some moments in time, or at some points in space. In this work, we explore the latter: we construct a linear array of demes in space, with local mating and competition, and occasional migration between neighboring demes. As the deme number increases, we consider performance on a progressively longer fraction of the entire walking task.

In Figure 5, demes k and k+1 are shown. In deme k, the population proceeds, much as before, from label 1, past label 2, to label 3. However, in this case, we have clipped the fitness function starting near label 3, such that any progress past a certain point receives the same score. We have created a neutral "mesa" on which the population is free to move by drift. (In practice, this translates, for example, to not caring how badly the robot flails its legs past a certain point in the test, but only scoring for performance before that point.) Only a single dimension of "progress" is shown here, but neutral drift will in practice entail movement in a high dimensional space. By Gavrilets' reasoning [10], multidimensional neutral networks can be far reaching, hence we can expect that some members of the population will eventually drift multi-dimensionally "around" label 3, which had been a local optimum in the previous figure, reaching label 4. (The distance from label 3 to 4 is not to scale; nor is any other distance in this cartoon schematic.) At label 4, some individuals, by chance, migrate (label 5) to deme k+1 (label 6). In deme k+1, more of the fitness function is exposed; the population can again clearly "see" a gradient toward higher fitness, which "leads" it forward to labels 7 and 8.

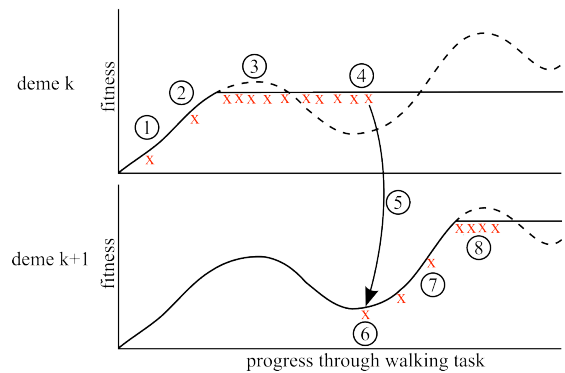

**Figure 5: Multiple demes: crossing a "fitness valley"**

We implemented this idea in practice on the robot as follows: We constructed 19 demes, numbered 0 through 18. A run begins with only deme 0 populated with individuals. In deme 0, we run the first 0.4 seconds of the walking test under neural control; then inhibit neural control for a further 0.6 seconds, terminating the test at a total length of 1.0 seconds. In the next deme (deme 1), we increase the length of the neural control phase from 0.4 to 0.6 seconds, then again run with neural control inhibited for 0.6 seconds, for a total test length (after walking is initiated) of 1.2 seconds. Each successive deme extends the controlled part of the test by 0.2 seconds, and maintains a 0.6 second period with neural control inhibited. By the 16th deme, the entire walking cycle has been tested. At this point we do not inhibit neural control; instead, we hand control back to the balancing network. The final 3 demes extend the period of balance control by .6 seconds each, for a total of 19 demes, numbered 0 through 18.

A detail: the sequence of demes does not map one-to-one to the sequence of neural networks. Generally, each deme adds 2 networks to the test, but during the three full steps in the middle of the walking sequence the same six networks are used three times (connected in topologies *walk-right*, *walk-left*, *walk-right*, respectively, as described above). This allows exploitation of the bilateral symmetry of the robot.

By inhibiting neural control at the end of the test, and stopping the test after 0.6 seconds, we have a period during which the behavior of the robot is determined by how successful the networks were in keeping the robot in a stable, upright position, and in a dynamical state requiring a minimum of future corrections to continue a successful walking pattern. This scheme allows the first networks in the sequence to progress towards higher fitness, without requiring the later networks to be operational.

## 2.8 Genetic Algorithm

The genotype of each individual is split into 22 chromosomes; each chromosome holds the weights for one of the 22 neural networks. Pairs of individuals reproduce sexually, with segregation of chromosomes, but no recombination within chromosomes. With a chance of 10%, a given chromosome will receive a single point mutation, which is assigned to one of its network weights randomly.

Mutation of a single weight occurs by the following procedure: We pick a number $r$ uniformly distributed between -1.1 and 0.9. We multiply the existing weight by $f = 10^r$. This has the effect of multiplying or dividing the current weight by as much as approximately 10, with a bias towards zero. Next, with a 10% chance we multiply the weight by -1 to flip its sign. If the original weight was exactly zero, however, then instead of the above, we simply pick a new weight uniformly distributed between -0.1 and 0.1.

The life cycle consists of the following, in this order: scoring of phenotypes; sexual reproduction proportional to scaled score, with segregation of chromosomes; mutations applied to offspring; migration. We use a population of $N$=30 in each deme. Each deme has a minimum fitness threshold; any individual below that threshold does not reproduce. We use this to impede invasion of higher numbered demes. A total of approximately 120 individuals are permitted to live, allotted to high demes first, and moving downward; this means that there are generally four demes with any living individuals (excluding a few new migrants) at any time. Deme 0 is initially populated, and individuals gradually invade demes one by one as they meet a threshold fitness value that we specify for each deme. Only individuals within the same deme can mate; scoring is also relative within demes only.

Migration occurs by the following procedure: all individuals have a real number in the range [0, 19), representing their position along the line of demes. Divisions between demes occur at integer values. At each generation, each individual takes a random step of a length and direction normally distributed about zero with variance 0.3; hence all individuals follow a random walk. Whenever an individual's value crosses an integer threshold, it has "migrated" to another deme.

## 3. RESULTS

Figure 6 plots the average score in each deme, by generation (some demes are omitted, for clarity). The mapping from demes to the walking task is as follows: deme 0 is the "step off", a lean to the left so that the robot can subsequently take a step with the right leg. Demes 1-3 correspond to the first, accelerating step with the right leg. The robot is now at walking speed. Demes 4-6, 7-9, and 10-12 correspond to coasting steps with the left, right, and left legs, respectively. Demes 13-15 correspond to a decelerating step with the right leg. Demes 16-18 correspond to balancing in a standing position for an increasing length of time.

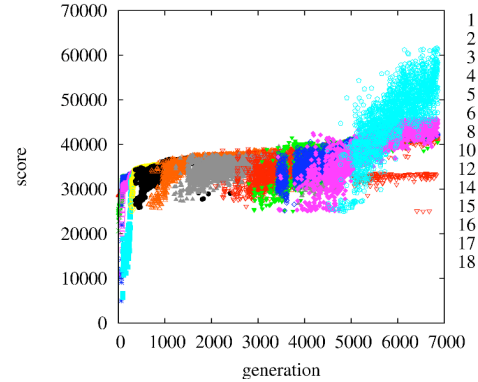

**Figure 6: Progress through the demes**

Evolution from deme 0 to the first invasion into deme 18 took approximately 4700 generations, and individuals in deme 18 continued to improve from there. Note that, because there are four active demes at a time, adaptive innovations from lower-numbered (but currently active) demes can still arise, and invade forward to the leading edge of progress, even after the first invasion of the higher-numbered deme. Therefore, it is not necessarily a waste of computation to continue to run individuals in lower-numbered demes, even though higher-numbered demes have already been invaded.

We ran multiple experiments with a single deme, i.e., without multiple demes and the progressive fitness functions described above. These runs all became trapped in local optima, and did not complete the walking task (before we judged them to be stuck, and terminated them), suggesting that our multi-deme progressive-fitness method did indeed permit the population to escape local optima and complete the walking task, for the fitness function as constructed.

Several time sequences of images of the simulated robot walking are shown in Figure 7; an animated movie is available here [9]. Each row corresponds to a sample run of the best-scoring individual from a certain deme at a certain generation (as labeled in the figure). In the first row, the best individual falls over almost immediately; as progress is made through the demes, the best individual improves until it completes the entire walking task.

## 4. CONCLUSIONS

We have presented elements of both engineering and theoretical interest. We describe a genetic representation scheme for a neural controller that was successfully evolved to produce dynamic walking in a simulated pneumatic (non-stiff) actuated biped. We have described a theoretical method to encourage populations to evolve "around" local optima. We are currently investigating other scoring functions to test the generality of this conclusion.

We have successfully run the evolved balancer on the hardware robot; a video is available here [7]. This success demonstrates a certain degree of simulator accuracy. We have not yet run the evolved walker described here on the hardware because we suspect that certain foot contacts with the floor are not accurately simulated. We plan to do further runs to evolve robustness to floor friction inaccuracy, hoping to attain a walk that does not rely on the particular errors in the simulation of floor friction made by the simulator. Specifically, the simulator is inaccurate if the robot drags its feet on the floor.

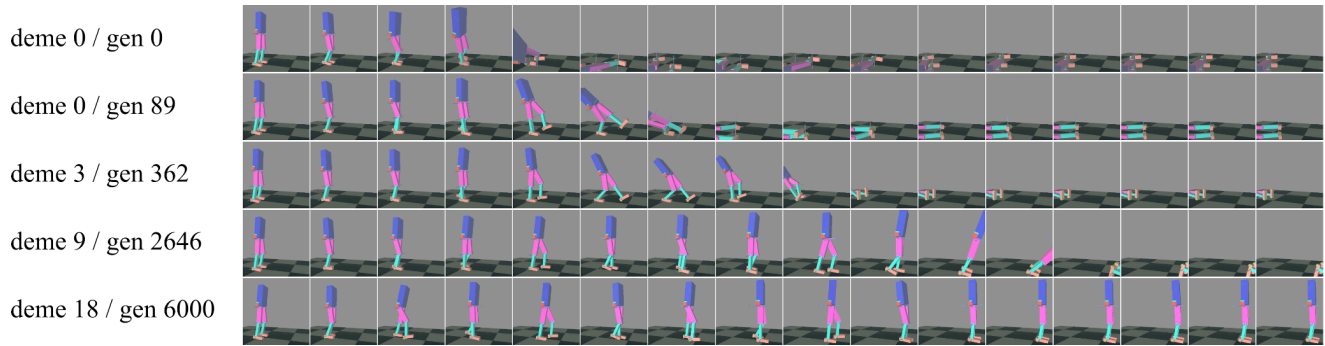

Figure 7: Performance on the walking task improves over time, as populations invade higher-numbered demes. An animated movie of this figure is available here [9].

Our multi-deme method does have a performance cost: in our case, we kept four demes full of individuals alive, rather than a single deme; this increases the per-generation cost linearly; however, this method solved the task, whereas the standard (single deme) method did not. Potentially, this is the key to solving much more difficult problems than are currently solvable with evolutionary computing. We believe that genetic representations with greater potential for evolvability (e.g., evolvable network topologies), when "encouraged" by means such as our multi-deme method, promise a way forward.

## 5. ACKNOWLEDGMENTS
The success of this project owes much to the work of the founder of Anybots, Trevor Blackwell. Thanks also to Elliot Cuzillo for his work on the physical simulator.